%% file: main.tex
\documentclass{article}
\usepackage[table]{xcolor}
\usepackage{spconf}
\usepackage{hyperref}
\usepackage{url}
\usepackage{times}
\usepackage{epsfig}
\usepackage{graphicx}
\usepackage{amsmath}
\usepackage{amssymb}
\usepackage{booktabs}
\usepackage{float}
\usepackage{algorithm}
\usepackage{algpseudocode}
\usepackage{subcaption}
\usepackage{nameref}
\usepackage{placeins}
\usepackage{enumitem}
\usepackage{mathtools}
\usepackage{verbatim}
\usepackage[british]{babel}

\DeclarePairedDelimiter\floor{\lfloor}{\rfloor}

\floatname{algorithm}{Pseudocode}

\DeclareMathOperator*{\argmin}{arg\,min}
\DeclarePairedDelimiter{\ceil}{\lceil}{\rceil}

\title{Deep Neural Networks Under Stress\thanks{\copyright 2016 IEEE --- Manuscript accepted at ICIP 2016.}}

\name{Micael Carvalho$^{(1,2)}$, Matthieu Cord$^{(1)}$, Sandra Avila$^{(2)}$, Nicolas Thome$^{(1)}$ \& Eduardo Valle$^{(2)}$}
\address{
    $(1)$ Sorbonne Universit\'es, UPMC Univ Paris 06, CNRS, LIP6 UMR 7606, 4 place Jussieu 75005 Paris\\
    $(2)$ University of Campinas, RECOD Lab -- DCA/FEEC/UNICAMP, Campinas, Brazil
}

\definecolor{Gray}{gray}{0.9}

\floatstyle{plain}
\newfloat{myalgo}{tbhp}{mya}
{\begin{myalgo}[#1]
\centering
\begin{minipage}{#2}
\begin{algorithm}[H]}%
{\end{algorithm}
\end{minipage}
\end{myalgo}}
\newcommand{\specialcell}[2][c]{\begin{tabular}[#1]{@{}c@{}}#2\end{tabular}}

\begin{document}
    \definecolor{Gray}{gray}{0.9}
    \maketitle
    
    \begin{abstract}
        In recent years, deep architectures have been used for transfer learning with state-of-the-art performance in many datasets. The properties of their features remain, however, largely unstudied under the transfer perspective.
        In this work, we present an extensive analysis of the resiliency of feature vectors extracted from deep models, with special focus on the trade-off between performance and compression rate. By introducing perturbations to image descriptions extracted from a deep convolutional neural network, we change their precision and number of dimensions, measuring how it affects the final score. We show that deep features are more robust to these disturbances when compared to classical approaches, achieving a compression rate of 98.4\%, while losing only 0.88\% of their original score for Pascal~VOC~2007.
    \end{abstract}
    
    \begin{keywords}
    feature robustness, deep learning, transfer learning, image classification, feature compression
    \end{keywords}
    
    \section{Introduction}
    \input{sections/01_introduction}

    \section{Transfer Strategies}\label{sec:proposed_approach}
    \input{sections/02_transfer_strategies}

    \subsection{Dimensionality Reduction (DR)}\label{sec:stress_1}
    \input{sections/02a_dimensionality}
    
    \subsection{Quantization (Q)}\label{sec:stress_2}
    \input{sections/02b_quantization}
    
    \subsection{Feature Compression (FC)}\label{sec:combined_stresses}
    \input{sections/02c_compression}
    
    \section{Experiments}\label{sec:internal_discussion}
    \input{sections/03_results}

    \section{Discussion}\label{sec:conclusion}
    \input{sections/04_conclusion}

    \vspace{-0.3cm}
    \section*{\normalsize ACKNOWLEDGEMENTS}\label{sec:acknowledgements}
    This research was partially supported by CNPq, Santander and %Project ``Capacita\c{c}\~ao em Tecnologia de Informa\c{c}\~ao" sponsored by
    Samsung Eletr\^onica da Amaz\^onia Ltda., in the framework of law No. 8,248/91. We also thank CENAPAD-SP (Project 533), Microsoft Azure and Amazon Web Services for computational resources and Michel Fornaciali for his valuable advices.
    
    \bibliographystyle{IEEEbib}
    \bibliography{references.bib}
\end{document}

%% file: sections/01_introduction.tex
Deep Convolutional Neural Networks have swept the Computer Vision community, with state-of-the-art performance for many tasks~\cite{Krizhevsky2012,He2015,Durand2016a}. However, an analytical understanding of their models is still lacking, shrouding their use under a cloud of ad hoc procedures --- tricks of the trade --- without which they simply fail to work. Therefore, a full understanding of deep representations became the new Holy Grail of research in Machine Learning and Computer Vision~\cite{Bruna2013, LeCun2015}.

We explore here the properties of Deep Networks, measuring to which extent they preserve discriminative information about the input, i.e., measuring the robustness of the feature vectors they generate. Indeed, we may understand a deep model as one that first learns to extract a good representation (feature extraction step) and then uses that representation to make a decision (classification or regression step). Most of the challenge in understanding deep models is due to the unknown nature of the learned features.

\begin{figure}[t]
    \begin{center}
    	\includegraphics[width=\columnwidth]{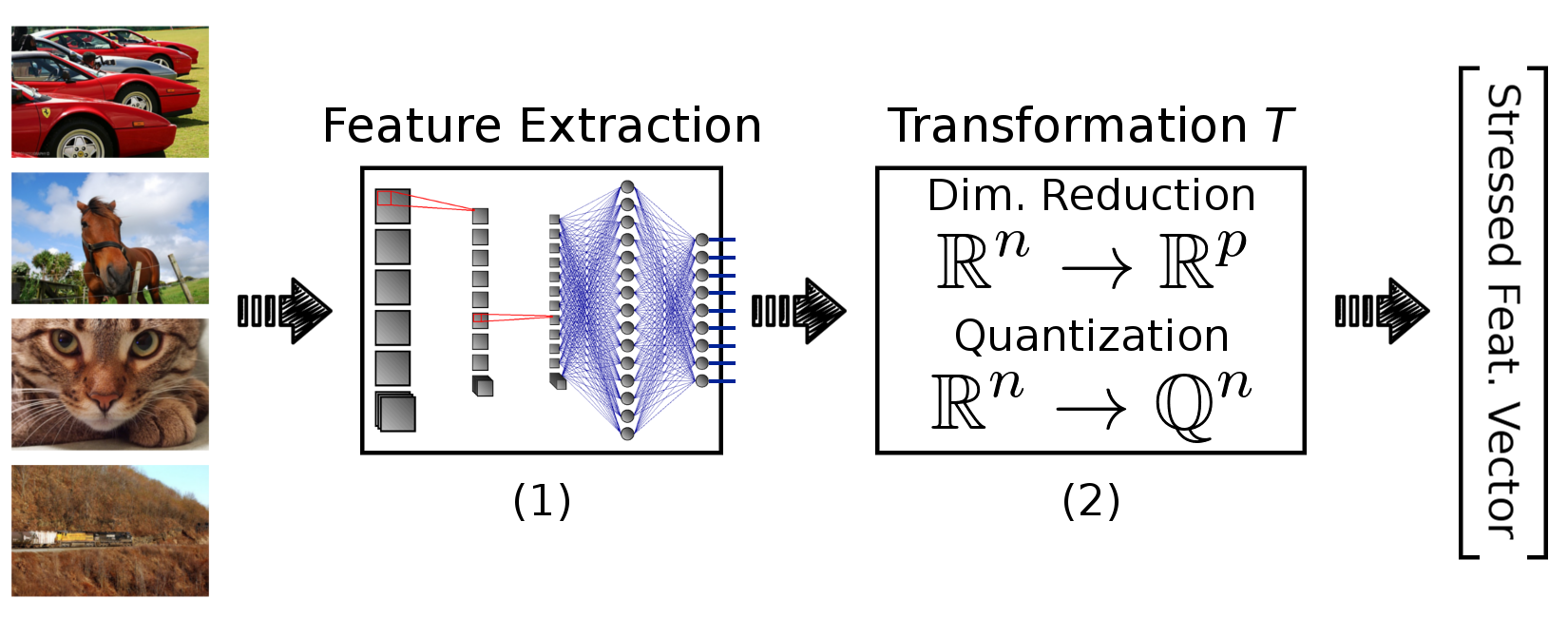}
    	\caption{Overview of our framework. Input images are converted to stressed feature vectors by: (1) extracting descriptions using a pre-trained deep network, (2) transforming/stressing the feature vectors by reducing their precision or their number of dimensions.\vspace{-0.2cm}}
    	\label{fig:scheme}
    \end{center}
\end{figure}

Pursuing that understanding, we use transfer learning and ``stress'' tests to probe the networks. Transfer learning consists in recycling knowledge from one model to another, in the form of model weights, initialization, or architecture (e.g. \cite{Razavian2014,Yosinski2014,Chevalier2015,DurandICCV15}), saving both computational resources and training data. Transfer learning is often used, with great success, on deep models, which are very greedy in terms of data and processing power. A straightforward scheme is to choose a pre-trained network, freeze the weights up to a certain layer, and to introduce and train new layers for the new task. By picking different layers from the original network, one controls the degree of transfer between the models. Conceptually, the output of the frozen transferred layers for any image may be seen as a feature vector $\mathbf{x}$. Thus, any classifier, like SVM, may be used for classification on a target dataset.

In this paper, we propose stress tests, represented in Figure~\ref{fig:scheme}, which consistently interfere in the network to selectively destroy information. We explore two important aspects of deep architectures: dimensionality and numerical precision of their representations. Dimensionality stress tests introduce $T:\mathbb{R}^{n}\rightarrow\mathbb{R}^{p}$, with $p$ smaller than the original dimensionality $n$. Quantization stress tests introduce $T:\mathbb{R}^{n}\rightarrow\mathbb{Q}^{n}$ where $\mathbb{Q}$ is a more aggressively quantized subset of real numbers than $\mathbb{R}$. We also combine the two stresses.

Although recent studies reevaluate deep architectures with respect to the size and precision of their representations (e.g. \cite{Vanhoucke2011,Courbariaux2014,Courbariaux2015,Judd2015}), their primary focus are practical impacts upon the original tasks. Our framework is designed for transfer learning tasks and, as we try to shed light on general properties of the networks, we will see that they show a strong degree of redundancy, opening the opportunity to create powerful compact descriptors.

%% file: sections/02_transfer_strategies.tex
Our main objective is to explore the VGG-M deep convolutional model \cite{Chatfield2014}, which was originally trained on ImageNet, in a transfer scheme for the classification task of the Pascal VOC~2007 dataset~\cite{Everingham10}. We perform extensive experiments to study the robustness of this architecture, detailed in Table~\ref{tab:matconvnet}, against different types of stresses.

Let us formalize the pre-trained deep model as a series of functions $\phi_i:\mathbb{R}^{m_i}\rightarrow \mathbb{R}^{n_i}$, where $\phi_i$ is the $i^{th}$ layer of the network, $m_1$~is equal to the dimensionality of the input data and $n_i =\nolinebreak m_{i+1}$ is the output of such layer.

In our stress tests, we choose a layer $i$ up to which we freeze the network (i.e., we keep layers $\phi_1...\phi_i$ untouched). At first, we use the output of layer $\phi_i$ to train an SVM. Then, we pick a stressing function $T$ and retrain the model using $T(\phi_i)$ as input. Comparing the two scores, we can infer the network's resiliency to the chosen stress.

To better highlight inherent properties of deep models, instead of specific characteristics of VGG-M~\cite{Chatfield2014}, we also evaluate part of our experiments with GoogLeNet~\cite{Szegedy2014}. Furthermore, in order to differentiate these deep models from classical approaches, we also report comparative results with the recent Bag-of-Words (BoW) model's BossaNova~\cite{Avila2013}. In all cases, we pre-process the images according to each model's recommended protocol. 

We also explore how to extend the results obtained for Pascal VOC~2007 by comparing part of the experiments with two other datasets: MIT-67~--~Indoor~\cite{Quattoni2009} and UPMC Food-101~\cite{XinWang2015} (67 and 101 classes, respectively).

%% file: sections/02a_dimensionality.tex
In order to understand how redundant is the deep representation, the first stress tests drop dimensions from the feature vector. The number of dimensions $p_i$ preserved at each step $1\leq\nolinebreak i \leq 20$ is proportional to the initial size $n$ of the feature vector, according to the expression $p_i=\floor{\dfrac{n*(21-i)}{20}}$.

We contrast two strategies for selecting the $p_{i-1} - p_{i}$ dimensions dropped at each step $i$: $\mathbf{T}_{\textbf{DR-1}}$ drops randomly; and $\mathbf{T}_{\textbf{DR-2}}$ uses a PCA-based strategy. The latter discards  the dimensions encoding less variance. To take in consideration the random choice in DR-1, we repeat the experiment 10 times.

\begin{table}[t]
    \begin{center}
    \noindent\makebox[\linewidth]{
        \resizebox{\columnwidth}{!}{
            \begin{tabular}{ccccccccc}
                \toprule
        		 & \textbf{G 1} & \textbf{G 2} & \textbf{G 3} & \textbf{G 4} & \textbf{G 5} & \textbf{G 6} & \textbf{G 7} & \textbf{G 8}\\
        		\toprule
        		\rowcolor{Gray}
        		\textbf{Conv.} & L 1 & L 5 & L 9 & L 11 & L 13 & -- & -- & -- \\
        		\textbf{Fully} & -- & -- & -- & -- & -- & L 16 & L 18 & L 20 \\
        		\rowcolor{Gray}
        		\textbf{ReLU} & L 2 & L 6 & L 10 & L 12 & L 14 & L 17 & L 19 & -- \\
        		\textbf{LRN} & L 3 & L 7 & -- & -- & -- & -- & -- & -- \\
        		\rowcolor{Gray}
        		\textbf{Pooling} & L 4 & L 8 & -- & -- & L 15 & -- & -- & -- \\
        		\textbf{Softmax} & -- & -- & -- & -- & -- & -- & -- & L 21 \\
        		\bottomrule
            \end{tabular}
        }
    }
    \end{center}
    \vspace{-0.2cm}
    \caption{VGG-M model. Description of layers (L) and groups (G) of the VGG-M model, from the MatConvNet toolbox~\cite{matconvnet}, proposed by Chatfield et al. \cite{Chatfield2014}. \textit{Conv.} indicates a convolutional layer, \textit{Fully} a fully connected layer, \textit{ReLU} a Rectified Linear Unit layer, \textit{LRN} a Local Response Normalization la\-yer, \textit{Pooling} a Max Pooling layer and \textit{Softmax} the activation of the Softmax~function.}
    \label{tab:matconvnet}
\end{table}

%% file: sections/02b_quantization.tex
The other stressor diminishes the numerical precision of the representation, quantizing the feature vectors. Our objective is not to explore advanced quantization strategies here, but to consider 2 fast and simple scalar quantizations and to analyze their effect on a classification task. In our first one, \textbf{Q-1}, all dimensions are quantized in the same $h \in [1,30]$ regular intervals, using the minimum ($min$) and maximum ($max$) scalar values observed in the training set for all dimensions. In our second one, \textbf{Q-2}, we adapt the limits for each dimension individually, according, again, to values observed in the training set.

Formally, Q-1, using the global step $ st = \frac{max-min}{h},$ has a single dictionary $\mathcal{H}$, generated by
\begin{gather*}
   ~~\mathcal{H} = \{(min + \frac{st}{2}) + st * i \mid~0 \leq i < h\}
\end{gather*}

For Q-2, let $\mathbf{x}$ be the feature matrix of the training feature vectors,  and $\mathbf{x}_t$ the $t^{th}$ element from all the vectors.
Using one step $ st_t = \frac{max(\mathbf{x}_t)-min(\mathbf{x}_t)}{h}$ per dimension, Q-2 has $n$ (number of dimensions) dictionaries, generated by
\begin{gather*}
    \mathcal{H}_t = \{(min(\mathbf{x}_t) + \frac{st_t}{2})+ st_t * i \mid~0 \leq i < h\}
\end{gather*}

Finally, in the quantization step, we assign to each element the value of the closest point in the dictionary. For Q-1 and Q-2, respectively, this is defined by:
\begin{gather*}
    T_{Q\text{-}1}(\mathbf{x}_{ij}) = \underset{y}{\argmin}~\{abs(\mathbf{x}_{ij} - y) \mid y \in \mathcal{H}\}\\
    T_{Q\text{-}2}(\mathbf{x}_{ij}) = \underset{y}{\argmin}~\{abs(\mathbf{x}_{ij} - y) \mid y \in \mathcal{H}_j\}
\end{gather*}

%% file: sections/02c_compression.tex
The final experiment \textbf{FC}, applies both stressors \mbox{DR-2} and \mbox{Q-2} simultaneously, dropping dimensions of the feature vector and quantizing the values of the remaining elements. Our goal is to measure any cross-effects between \mbox{DR-2} and \mbox{Q-2}.

%% file: sections/03_results.tex
As explained, for a given experimental point, we freeze a pre-trained network at layer $\phi_i$, discarding all upper layers. We then pick a stressing function $T$, and use the output of $T(\phi_i)$ as a feature vector in a transfer learning classification task. We $\ell_2$-normalize those feature vectors, and feed them to a linear SVM model\footnote{For all setups, we use a regularization parameter $C = 1$; preliminary experiments shown very little variation when the $C$ was cross-validated.}~\cite{liblinear}, measuring the model's scores for different choices of $T$. By picking stressing functions of different kinds and intensities (including the identity $T(x) =\nolinebreak x$) we gain insight on the resiliency of deep models to those~stresses.

For all our experiments, we report the classification scores in Mean Average Precision (mAP) for Pascal VOC~2007, and Accuracy (Acc) for Food-101 and MIT-67, following literature tradition on those datasets. 
Although we have tested the deep networks extensively, due to space constraints, we only report the experiments with layer~19 for the VGG-M, and with layer 151 for GoogLeNet. Those results are representative of our observations throughout the networks.

\begin{figure}[bt]
    \centering
	\includegraphics[width=0.85\columnwidth]{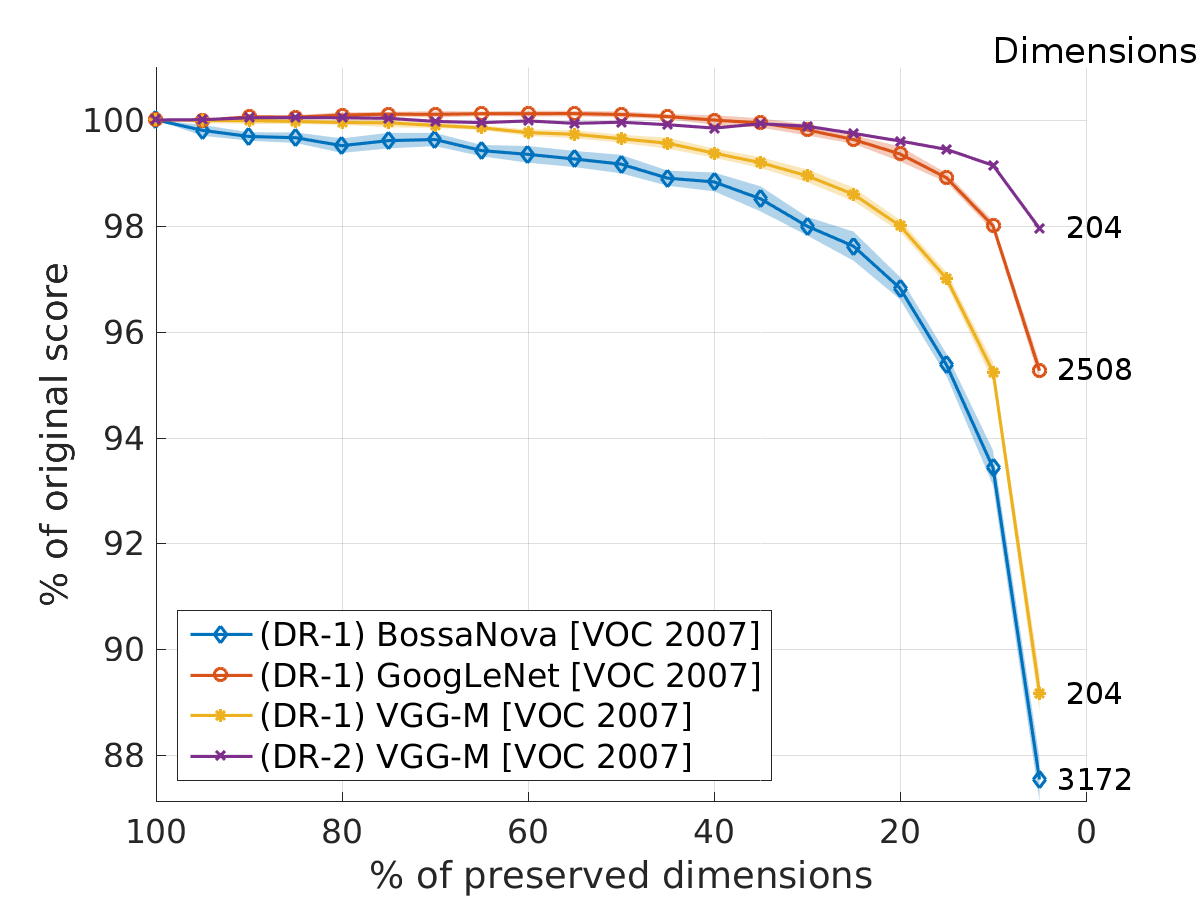}
    \caption{Results for dimensionality reduction (DR) on VOC~2007 with standard deviation (shaded regions around the lines). In the horizontal axis, each value indicates the percentage of the original dimensions that is kept, while the corresponding score, with the respect to the initial one, is shown vertically. On the right side of the figure, we show the number of dimensions for each model, when only 5\% of their initial size is preserved.}
	\label{fig:voc}
\end{figure}

\begin{table}[b]
    \begin{small}
    \begin{center}
      \noindent\makebox[\linewidth]{
        \resizebox{\columnwidth}{!}{
            \begin{tabular}{ccccc}
                \toprule
                & \textbf{VGG-M} & \textbf{GoogLeNet} & \textbf{BossaNova}\\
        		\toprule
        		\rowcolor{Gray}
        		Pascal VOC 2007 (mAP) & 76.95\% & 80.58\% & 51.02\% \\
        		MIT-67 Indoor (Acc) & 63.35\% & -- & -- \\
        		\rowcolor{Gray}
        		UPMC Food-101 (Acc) & 46.22\% & -- & -- \\
        		\midrule
        		Feature Dimensionality & $4*10^3$ & $5*10^4$ & $6*10^4$ \\
        		\bottomrule
            \end{tabular}
        }
    }
    \end{center}
    \vspace{-0.2cm}
    \caption{Classification scores for deep and BoW strategies in a vanilla transfer scheme, with a linear SVM as classifier.}
    \label{tab:baselines}  
    \end{small}
\end{table}

Table~\ref{tab:baselines} shows the scores for our vanilla experiments, using setups without perturbating the feature vectors (i.e., $T(x) = x$). We simplify BossaNova's pipeline for Pascal VOC~2007, disabling the concatenation with the classic Bag of Visual Words, and using a linear SVM instead of the recommended RBF kernel.

The results for our dimensionality reduction (DR) experiments in the Pascal VOC~2007 dataset are shown in Figure~\ref{fig:voc}. Strong redundancy on the representations is detected, since across runs, small variations in the score were observed. GoogLeNet was the most robust against random dimensionality perturbation DR-1, with an average mAP drop of 4.74\% for 95\% of the dimensions removed. However, GoogLeNet is, from start, 12-times bigger than CNN-M. Considering a direct comparison of descriptions of approximately the same size, the scores of the two models were equivalent.

Although BossaNova has shown similar resiliency to dimensionality reduction, VGG-M held better scores for every test point, despite having feature vectors 15-times smaller (right side of Figure~\ref{fig:voc}).

The PCA-based strategy was very effective for preserving information while dropping dimensions. DR-2 held 97.95\% of the mAP with 95\% of dimensions removed, while DR-1 could only keep 89.16\% of the mAP. Choosing the right dimensions to drop improves the robustness of the feature vectors to dimensionality perturbations.

\begin{figure}[tb]
    \centering
	\includegraphics[width=0.85\columnwidth]{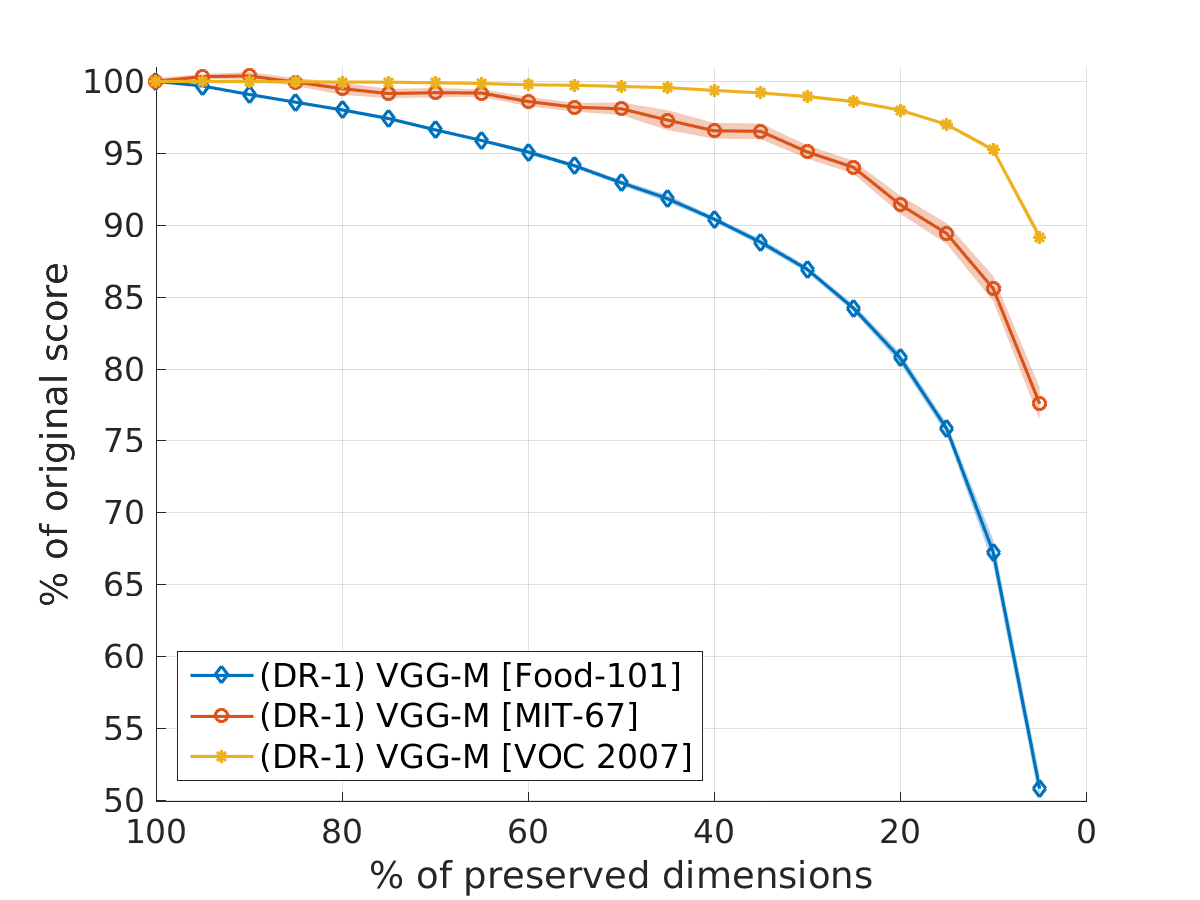}
    \caption{Results for dimensionality reduction (DR) with \mbox{VGG-M} for Food-101, MIT-67 and VOC~2007. The datasets have 101, 67 and 20 classes, respectively.}
	\label{fig:vgg}
\end{figure}

The number of classes in the target dataset also seems to play an important role on performance resiliency, as seen in Figure~\ref{fig:vgg}. For correctly classifying the data, diverse datasets may need complementary feature points, which can be lost with dimensionality reduction.

\begin{figure}[ht]
    \begin{center}
        \centering
    	\includegraphics[width=0.85\columnwidth]{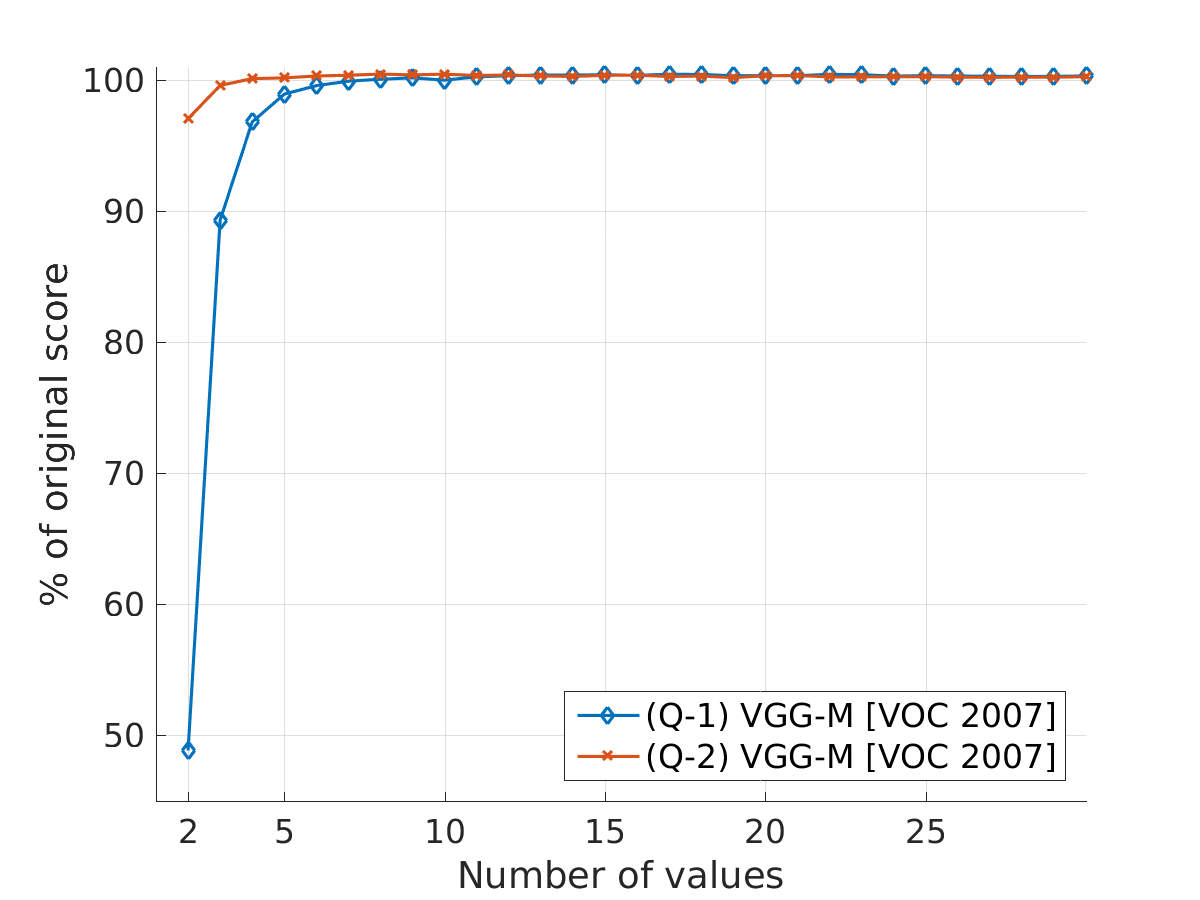}
    	\caption{Results for quantization of features on the base setup (VGG-M and Pascal VOC~2007). We can keep vanilla performance while reducing the feature vectors from $32*m_i$ to $\ceil{\log_2 7}*m_i$ and $\ceil{\log_2 4}*m_i$ bits, using Q-1 and Q-2, respectively.}
    	\label{fig:quantization}
    \end{center}
\end{figure}

Our quantization (Q) experiments, on the other hand, reduce the size of the feature vectors, from initial $32 * m_i$ bits\footnote{For 32-bit single-precision floating-point numbers.}, by aggressively limiting their values. Q-2 performed better than Q-1, indicating that adaptiveness to scale plays an important role (Figure~\ref{fig:quantization}).

\begin{table*}[tb]
    \begin{small}
    \begin{center}
    %\noindent\makebox[\linewidth]{
        %\resizebox{\columnwidth}{!}{
            \begin{tabular}{ccccccccc}
                \toprule
                & \textbf{\specialcell{Original Score}} & \textbf{\specialcell{DR-1 -- 2\%}} & \textbf{\specialcell{DR-1 -- 5\%}} & \textbf{\specialcell{DR-2 -- 1\%}} & \textbf{\specialcell{Q-1 -- 1\%}} & \textbf{\specialcell{Q-1 -- 4\%}} & \textbf{\specialcell{Q-2 -- 1\%}} & \textbf{\specialcell{Q-2 -- 3\%}}\\
        		\toprule
        		\rowcolor{Gray}
        		VGG-M & 76.95\% & 25\% & 10\% & 10\% & 6 values & 4 values & 3 values & 2 values \\
        		GoogLeNet & 80.58\% & 10\% & 5\% & -- & -- & -- & -- & -- \\
        		\rowcolor{Gray}
        		BossaNova & 39.59\% & 50\% & 25\% & -- & -- & -- & -- & -- \\
        		\bottomrule
            \end{tabular}
        %}
    %}
    \end{center}
    \vspace{-0.4cm}
    \caption{Minimum representation rate for Pascal VOC~2007. Each column indicates a requirement, and each line represents a dataset. The cells reveal the minimum representation needed for losing at most the indicated percentage. For instance, (DR-1~--~2\%) + GoogLeNet = 10\% means that with only 10\% of the dimensions, we lose at most 2\% of the initial~score.}
    \label{tab:summary}
    \end{small}
\end{table*}

Furthermore, Q-1 kept vanilla scores with 7 values, while Q-2 only needed 4. That represents a strong compression of the feature vectors, from $32*m$ to $\ceil{\log_2 4}*m$~bits.

The main results for DR and Q for VOC~2007 are summarized in Table~\ref{tab:summary}, where each column indicates the maximum desired loss with respect to the original score for an experiment, while the cells indicate the minimum value which satisfies such requirement. For example, the second line of the second column reveals that with only 10\% of the dimensions preserved, GoogLeNet score drops less than 2\%.

\begin{figure}[tb]
    \begin{center}
    	\includegraphics[width=0.9\columnwidth]{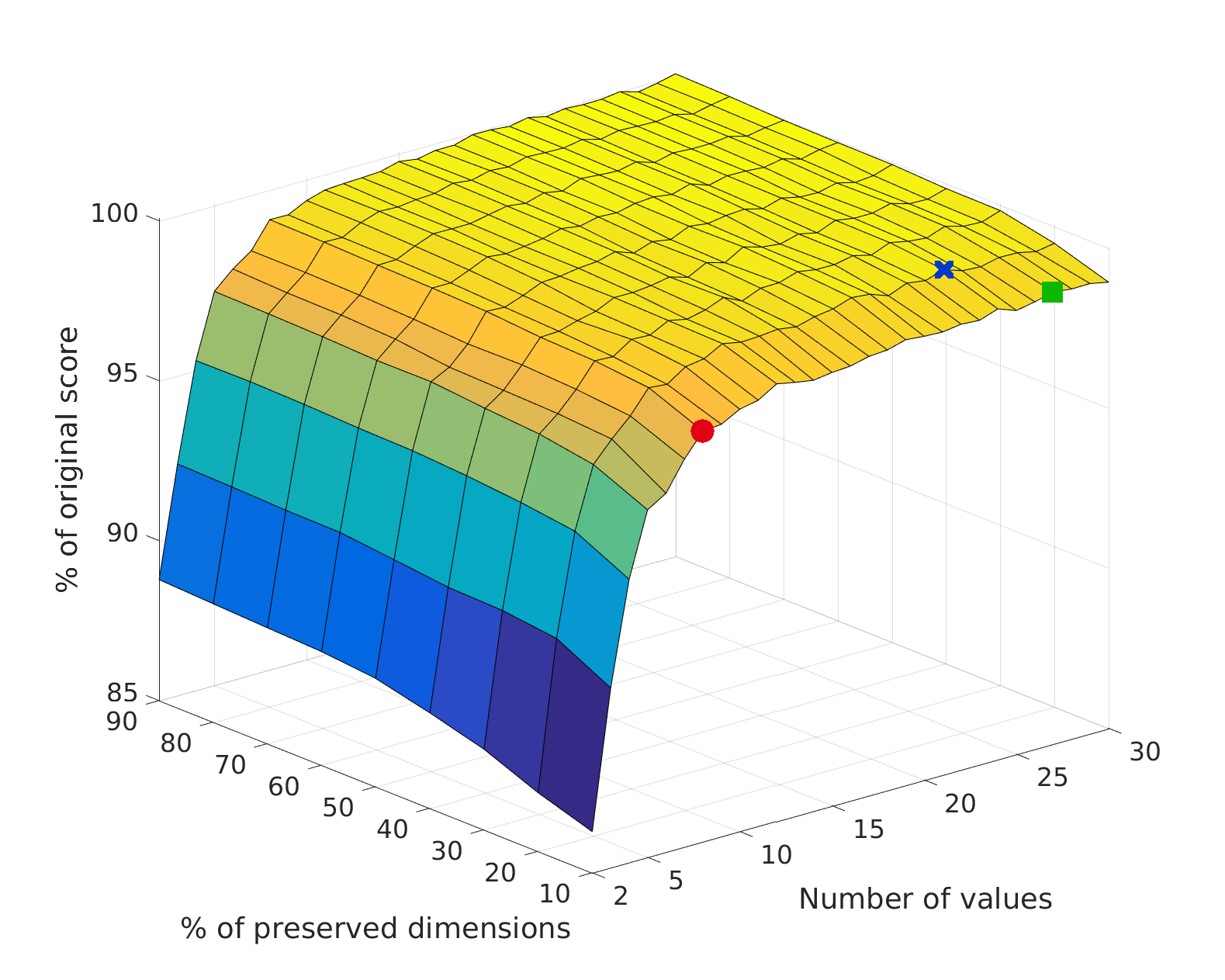}
    	\caption{Results for feature compression (FC). We reduce the number of dimensions and the precision of the feature vectors at the same time. The circle, square and cross, mark configurations with compression rates of 99.1\%, 98.4\% and 96.9\%, respectively, while maintaining 97.8\%, 99.1\% and 99.6\% of the original score.}
    	\label{fig:compression}
        \vspace{-1em}
    \end{center}
\end{figure}

Finally, the results for FC with the base setup are shown in Figure~\ref{fig:compression}. The flat region on the top represents combinations of parameters from DR-2 and Q-2 with complementary characteristics, indicating that the features can be compressed in terms of dimension and precision at the same time. We point, with the circle, square and cross markers, specific combinations of DR-2 and Q-2 with compression rates of 99.1\%, 98.4\% and 96.9\%, respectively, while maintaining 97.8\%, 99.1\% and 99.6\% of the original score.

Supplementary results and resources, including the source code for our experiments, are available online\footnote{{\url{https://github.com/MicaelCarvalho/DNNsUnderStress}}}.

%% file: sections/04_conclusion.tex
In this paper we evaluated the robustness of deep representations by introducing perturbations to feature vectors extracted from upper layers of deep networks. We explored in depth the resiliency of features transferred from the VGG-M model to the Pascal~VOC~2007 dataset. Our findings show that there is a high level of redundancy in deep representations, and thus, they may be heavily compressed. In our experiments, we achieve a compression rate of 98.4\%, while losing only 0.88\% of the original score for Pascal~VOC~2007. To ensure our conclusions are not dataset- nor model-specific, our two main approaches -- Dimensionality Reduction and Quantization -- were extensively tested, with supplementary results for MIT-67, Food-101, GoogLeNet and BossaNova. Furthermore, we observed that despite being more compact, deep architectures are also more robust to perturbations, when compared to approaches based on Bags of Visual Words. Those findings are specially useful for image retrieval and metric learning~\cite{LeBarz2015}, in which the size of the feature vector is crucial to achieve fast response times, and for applications involving portable devices or remote classification, in which data must be efficiently transferred over the~network.